\documentclass[letterpaper, 10 pt, conference]{ieeeconf}  
\usepackage{graphicx}
\usepackage{amsmath}
\usepackage{pgfplots}
\newcommand{\argmin}{\mathop{\rm arg~min}\limits}

\IEEEoverridecommandlockouts                              

\title{\LARGE \bf StereoVAE: A lightweight stereo-matching system using embedded GPUs}

\author{Qiong Chang$^{1*}$, Xiang Li$^{2}$, Xin Xu$^{2}$, Xin Liu$^{3}$, Yun Li$^{2}$ and Jun Miyazaki$^{1}$
\thanks{$^{1}$ Qiong Chang and Jun Miyazaki are with the School of Computing, Tokyo Institute of Technology, Tokyo, 152-8550, Japan.          
Qiong Chang is the corresponding author. {\tt\small q.chang@c.titech.ac.jp}
         }%
\thanks{$^{2}$ Xiang Li, Xin Xu and Yun Li are with the School of Electronic Science \& Engineering, Nanjing University, Nanjing, 210093, China.}
\thanks{$^{3}$ Xin Liu is with the Artificial Intelligence Research Center,
  National Institute of Advanced Industrial Science and Technology (AIST),
  Tokyo, 135-0064, Japan.}
}
\begin{document}

\maketitle
\thispagestyle{empty}
\pagestyle{empty}

\begin{abstract}
We propose a lightweight system for stereo-matching using embedded graphic processing units (GPUs).
The proposed system overcomes the trade-off between accuracy and processing speed in stereo
matching, thus further improving the matching accuracy while ensuring real-time processing.
The basic idea is to construct a tiny neural network based on a
variational autoencoder (VAE) to achieve the upscaling and refinement a small size of coarse
disparity map. This map is initially generated using a traditional matching method.
The proposed hybrid structure maintains the advantage of low computational complexity found in traditional
methods. Additionally, it achieves matching accuracy with the help of a neural network.
Extensive experiments on the KITTI 2015 benchmark dataset demonstrate that our tiny
system exhibits high robustness in improving the accuracy of coarse
disparity maps generated by different algorithms, while running in real-time on embedded GPUs. \textit{The code has been published on: https://github.com/changqiong/sRRNet}
\end{abstract}

\section{INTRODUCTION}
Stereo-matching is the task of measuring the distance of pixels relative to a camera. 
The depth information of an object or scene is significantly important in many fields including robotic vision, 3D reconstruction, driver
assistance systems etc.
It can be extracted from stereo-image pairs using pixel matching along an epipolar line.
Although matching methods are more susceptible to a lighting
environment than radar and structured-light technologies, they are capable of easily extracting more information (e.g. the shape feature). Thus, they can be efficiently used in practical applications, such as recognition and reconstruction, in a more flexible manner and at a lower cost than conventional technologies.
Recent stereo-matching methods can be divided into the following two categories: traditional
methods and learning-based methods. The main difference between these two categories lies in the calculation of the matching costs for each pixel, which is the most critical step in matching. 
Traditional methods, such as Census and normalized cross-correlation
(NCC), usually define a feature template for each pixel to complete the
matching. In these methods, the computational cost is low, but only limited accuracy can be achieved. On
the other hand, learning-based methods can extract more complex features by training
a neural network~\cite{c1}. This network provides high accuracy but has a high computational cost. 
Due to their massive advantage regarding the
amount of calculations, most current embedded systems often employ traditional
methods to estimate the depth information~\cite{c2}~\cite{c3}. However, this leads to far
less accuracy than that achieved by systems based on general-purpose platforms and often
requires auxiliary equipment.
To achieve improved accuracy, many studies have recently focused on implementing
lightweight convolutional neural networks (CNN) for embedded
graphic processing units (GPUs)~\cite{c4}-\cite{c7}. However, since most current embedded
systems require a significant compromise in terms of processing speed, they often prune
their neural networks as much as possible, resulting in limited accuracy.
Current CNN-based embedded stereo-vision systems exhibit a limited accuracy improvement over traditional methods in real-time processing~\cite{c2}. Thus, traditional methods are not far
behind in extracting matching features.

\begin{figure}
\center
\begin{tikzpicture}[scale=0.85]
\hspace*{-0.3cm}
\begin{axis}
[xlabel= Error Rate ($\%$) in KITTI 2015 Benchmark, 
ylabel= Time ({\itshape ms}),
height=2in,
width=4in,
legend style={font=\scriptsize},
legend entries ={AnyNet\cite{c4}, RTSMNET\cite{c8}, GPUBNN\cite{c9}, ReS2tAC\cite{c3}, SGMGPU\cite{c10}, Z2ZNCC\cite{c2}, \bf{Our StereoVAE}}, 
legend pos = north east,
ymax = 110,
xmax = 16,
xtick={0,2,...,16},
yticklabels={10,30,50,70,90,110},
ytick={10,30,50,70,90,110,130}
] 
\addplot[color=cyan!40!gray,mark=+,mark size=4pt, line width=0.5pt] coordinates 
{ 
  (	9.7	,	16.58	)

};

\addplot[color=cyan!60!gray,mark=asterisk,mark size=4pt] coordinates 
{ 
  (	3.57	,	87.7	)
};
\addplot[color=cyan!60!gray,mark=star,mark size=4pt] coordinates 
{ 
  (	4.57	,	100	)
};
\addplot[color=cyan!60!gray,mark=10-pointed star,mark size=4pt] coordinates 
{ 
  (	6.65	,	66	)
};

\addplot[color=cyan!60!gray,mark=otimes*,mark size=3.5pt] coordinates 
{ 
  (	8.24	,	8.93	)
};

\addplot[color=cyan!60!gray,mark=square*,mark size=3.5pt] coordinates 
{ 
  (	7.73	,	12.64	)
};

\addplot[color=red!60!white,mark=diamond*,mark size=4pt] coordinates 
{
  (	5.24	,	32.5	)
};

\end{axis}
\end{tikzpicture}
\caption{Accuracy comparison with other systems implemented on a Jetson AGX Xavier GPU.}
\label{fig:comparison}
\end{figure}
Based on the above, we propose a fast stereo-matching frame that combines the
advantages of traditional and CNN-based methods as follows:
\begin{itemize}
\item we first use the zero-mean normalized
  cross-correlation (ZNCC) and semi-global matching (SGM) of traditional methods to perform the
  essential stereo matching in a low-resolution image pair and generate an
  initial disparity map. Then, we enlarge and optimize this map using a neural network based on a variational
  autoencoder (VAE);
\item we propose a tiny VAE-based super-resolution network (StereoVAE),
  which employs a quarter size of the initial disparity map as an input and performs
  the upscaling and refinement;
\item we implement our real-time stereo-vision system on a Jetson AGX Xavier
  GPU and achieve a 5.24\% error rate at 30fps by employing the KITTI 2015 dataset.
\end{itemize}
The proposed hybrid structure can significantly reduce the running time by exploiting
the low computational cost of traditional methods and small images. It then uses the VAE-based neural network, which performs the upscaling and refinement, to compensate for the loss of accuracy.
Figure~\ref{fig:comparison} shows the performance comparison of the proposed system with
other embedded stereo-vision systems considering the KITTI 2015 benchmark dataset. It is observed
that our system achieves the best performance (close to the origin point), low running time (32.5ms) and low error rate (5.24\%).
The remainder of this paper is organized as follows. Section~\ref{sec:2}
describes the related work. Section~\ref{sec:3} presents the generation methods
for a low-resolution disparity image and StereoVAE structure. The
experimental results are discussed in section~\ref{sec:4}. Section~\ref{sec:5}
summarizes our work and presents a future research direction.

\section{Related Work}
\label{sec:2}
This section reviews some fast stereo-vision systems developed using GPUs. All
these systems employ neural networks to perform the matching and aim at achieving a 
high real-time processing speed by employing the KITTI 2015 dataset~\cite{c11}.

Duggal et al.~\cite{c12} proposed a method for pruning neural networks in a
GPU. They employed a CNN method to extract matching features and then a pruning
module based on PatchMatch~\cite{c13} to reduce the amount of calculation using cost aggregation. Zhang et al.~\cite{c14} proposed a cost aggregation
method to reduce the memory and computational cost in a GPU. They added a locally
guided aggregation layer, where three K$\times$K filters were employed for each
pixel to reduce the loss caused by the downsampling and upsampling layers. Both the above systems achieved high accuracy ($<3\%$ error rate) by employing the KITTI 2015 dataset but
low processing speed even on high-end GPUs ($<20$ fps).
Wang et al.~\cite{c4} proposed a four-stage lightweight neural network based
on a Jetson AGX Xavier GPU. First, they obtained only 1/16
of the original size of the disparity map in the initial stage and then propagated the intermediate
result to the next stage using a residual network block. This pyramid structure
was used to extract the features of different scales and effectively improved the matching
accuracy. Their system was the first to achieve error rates of 6.2\%--14\% at 26--82 fps,
making it the fastest learning-based stereo-vision system known in embedded platforms.
Chang et al.~\cite{c6} constructed a pyramid network similar to the
AnyNet~\cite{c4} and introduced a new attention-aware feature aggregation
module. This novel module effectively improved the
representational capacity of features without the need for excessive additional
calculations. Their system was able to perform stereo matching with an
1242$\times$375 image at 12--33 fps on a Jetson Tx2 module and achieved a 7.54\% of error
rate.
To further improve the matching accuracy, Gan et al.~\cite{c5} developed
a self-adaptive network with an extra convolutional spatial propagation network
to refine a coarse disparity map. Using this propagation network,
their system reduces the error rate to less than 5\% by employing the KITTI 2015
dataset. Furthermore, Dovesi et al.~\cite{c7} adopted a semantic
segmentation structure to further reduce the error rate to 3.3\%. However, both the above networks come at a speed price, taking more than 10 ms to process an 1240$\times$375 image.
In addition to the above-mentioned end-to-end systems, the matching accuracy also was improved in~\cite{c15} to~\cite{c17} using neural networks to optimize the disparity maps. However, these networks are too
large for embedded GPUs with limited resources and cannot meet the high-speed
processing requirements of embedded applications.
\section{Proposed System}
\label{sec:3}
In this section, we introduce the architecture of the proposed system, including 1) the
traditional Census and SGM methods for generating low-resolution raw
disparity maps and 2) the StereoVAE structure for upscaling and
refining the generated low-resolution disparity maps.

\subsection{Coarse Disparity Map Generation}
\subsubsection{Zero-means Normalized Cross-Correlation}
ZNCC is a template method that used to apply a matching as follows:
\begin{equation}
   C_{ZNCC}(x,y,d) = 1-\frac{\sum \limits_{(x,y)\in W}{\Delta I_R(x, y) \cdot
       \Delta I_T(x-d,y)}}{\sigma_R(x,y) \cdot \sigma_T(x-d, y)}, 
\label{equ:ZNCC}
\end{equation}
where
\begin{equation} 
\begin{split}
  \sigma_R(x,y) &= \sqrt{\sum \limits_{(x,y)\in W}\Delta I_R(x,y)^2},\\
  \sigma_T(x-d,y)&=\sqrt{\sum \limits_{(x,y)\in W} \Delta I_T(x-d, y)^2},\\
\end{split}
\end{equation}
and
\begin{equation}
\begin{split}
  \Delta I_R(x,y) &= I_R(x,y) - \overline{I_R(x,y)},\\ 
  \Delta I_T(x-d, y) &= I_T(x-d,y)-\overline{I_T(x-d, y)}.\\
\end{split}
\end{equation}
Here, the value of $C_{ZNCC}(x, y, d)$ in Eq.~\ref{equ:ZNCC}
represents the similarity between pixels $I_R(x,y)$ and $I_T(x-d,y)$,
which is within the range of [0,1]. When the value is small, the similarity is high.
In addition, $\sigma_R(x,y)$ and $\sigma_T(x-d, y)$ are the standard deviations of the pixel
values in window $W$ and they are used to normalize the correlation coefficient
between them.
\subsubsection{Semi-Global Matching}
SGM is one of the most popular optimization methods used for stereo
matching~\cite{c18}. It applies dynamic programming by treating the different
path directions equally.
The corresponding formula is:
\begin{equation}
  \begin{aligned}
    C_r(x,y,d) = &C(x,y,d) + \min(C_r(x-r,y,d),\\ 
                 &C_r(x-r,y,d-1)+P1,\\ 
                 &C_r(x-r,y,d+1)+P1,\\ 
                 &\min \limits_{i}C_r(x-r,y,i)+P2)\\
                 -&\min\limits_{k}C_r(x-r,y,k). 
   \label{equ:SGM}
  \end{aligned}
\end{equation}
Here, $C_r(x,y,d)$ represents the optimized cost along the $r$ direction, and
$C(x,y,d)$ represents the matching cost between pixels $I_R(x,y)$ and
$I_T(x-d,y)$. $P1$ and $P2$ are penalty constants for disparity
changes, and $i$ represents the disparity values, except for $d$ and $d\pm{1}$.
By adding the minimum cost value of the previous pixel with $P1$ and $P2$, the
effect of the adjacent pixels is propagated, whereas subtracting the minimum
cost value of the previous pixel ensures that the cost value does not overflow.
The final cost values $C_{SGM}(x,y,d)$ are then aggregated using different
directions as follows:
\begin{equation}
  C_{SGM}(x,y,d) = \sum \limits_{r}C_r(x,y,d) 
\label{equ:SGM1}
\end{equation}

\begin{figure*}
\centering \includegraphics[width=6.5in]{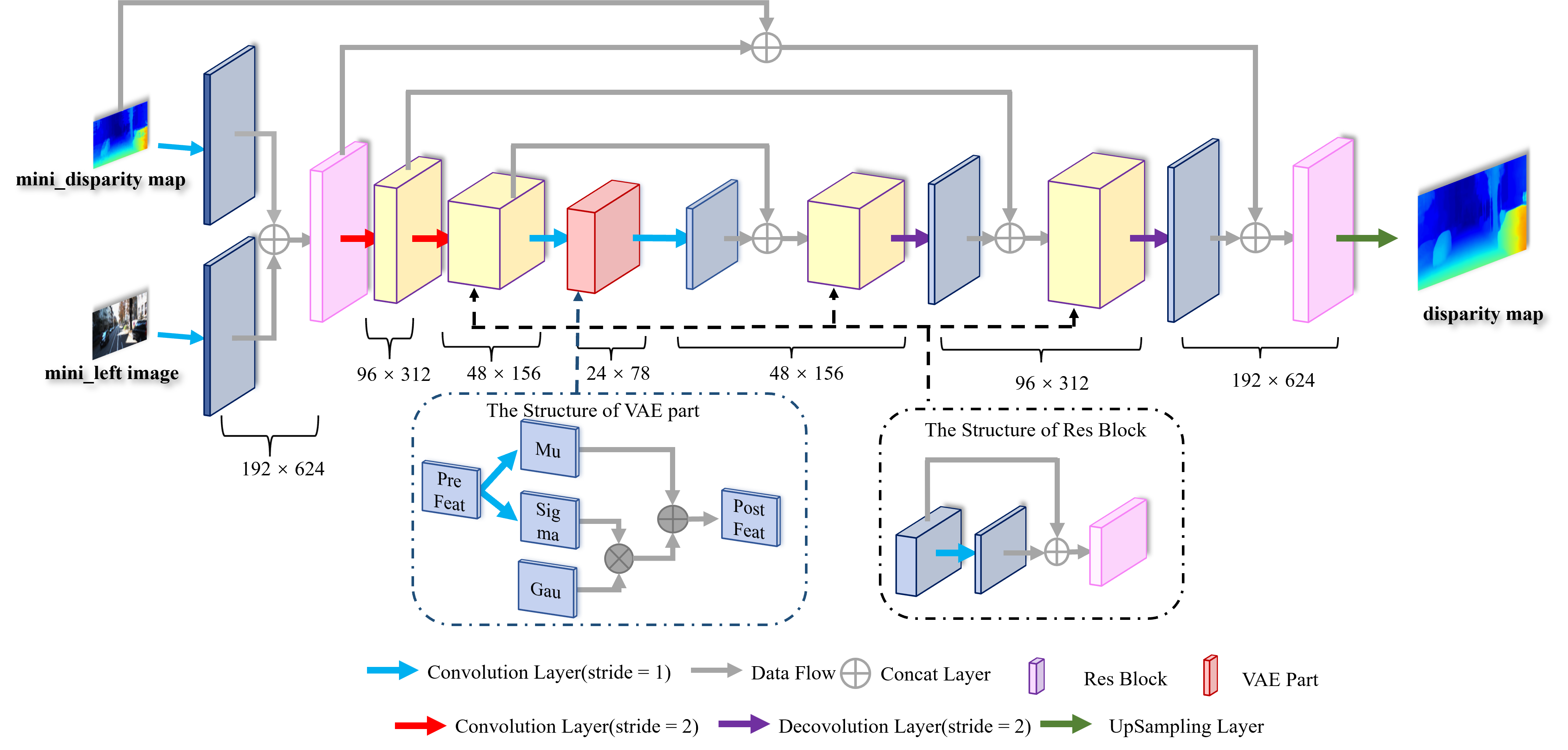}
\caption{StereoVAE Structure}
\label{fig:structure}
\end{figure*}
\subsubsection{WTA and Post-processing}
Based on the aggregation results, the disparity map can be generated using
\begin{equation}
  D_{map}(x,y) = \argmin_{d}{(C_{Final}(x,y,d))},
  \label{equ:WTA1}
\end{equation}
where $C_{Final}(x,y,d)$ represents the final matching costs aggregated using
Eq.\ref{equ:SGM1}.
Then, we use the single matching phase (SMP) method~\cite{c19} to remove the occlusion
parts and a median filter to reduce the amount of noise.
According to the SMP methods, no pixels in the reference image can be matched
by two or more pixels in the target image at the same time.
Therefore, only the values that meet the following conditions are valid:
\begin{equation}
 \forall{k\in(0,D)}:|x-D_{map}(x,y)-(x-k-D_{map}(x-k,y))|\leq{1}.
\end{equation}
The invalid values are replaced by the closest valid values.

\subsection{Disparity Upscaling and Refinement}
\subsubsection{StereoVAE structure}
\begin{table*}[t]
\centering
\caption{STEREOVAE STRUCTURE}
\label{STRUCTURE OF STEREOVAE}
\begin{tabular}{ccccccc}
\hline
\multicolumn{7}{c}{FEATURE EXTRACTION} \\ \hline
\multicolumn{1}{c|}{Layer} &
  \multicolumn{1}{c|}{Input} &
  \multicolumn{1}{c|}{Input Size} &
  \multicolumn{1}{c|}{Output} &
  \multicolumn{1}{c|}{Output size} &
  \multicolumn{1}{c|}{Output Kernel Number} &
  Kernel Size \\ \hline
\multicolumn{1}{c|}{1} &
  \multicolumn{1}{c|}{X\_0} &
  \multicolumn{1}{c|}{H/2 * W/2} &
  \multicolumn{1}{c|}{FE\_1} &
  \multicolumn{1}{c|}{H/2 * W/2} &
  \multicolumn{1}{c|}{32} &
  5 * 5 \\ \hline
\multicolumn{1}{c|}{1} &
  \multicolumn{1}{c|}{X\_1} &
  \multicolumn{1}{c|}{H/2 * W/2} &
  \multicolumn{1}{c|}{FE\_2} &
  \multicolumn{1}{c|}{H/2 * W/2} &
  \multicolumn{1}{c|}{32} &
  5 * 5 \\ \hline
\multicolumn{7}{c}{ENCODER} \\ \hline
\multicolumn{1}{c|}{Layer} &
  \multicolumn{1}{c|}{Input} &
  \multicolumn{1}{c|}{Input Size} &
  \multicolumn{1}{c|}{Output} &
  \multicolumn{1}{c|}{Output size} &
  \multicolumn{1}{c|}{output kernel number} &
  Kernel Size \\ \hline
\multicolumn{1}{c|}{1} &
  \multicolumn{1}{c|}{FE\_1 \& FE\_2} &
  \multicolumn{1}{c|}{H/2 * W/2} &
  \multicolumn{1}{c|}{EN\_1} &
  \multicolumn{1}{c|}{H/4 * W/4} &
  \multicolumn{1}{c|}{16} &
  5 * 5 \\ \hline
\multicolumn{1}{c|}{1} &
  \multicolumn{1}{c|}{EN\_1} &
  \multicolumn{1}{c|}{H/4 * W/4} &
  \multicolumn{1}{c|}{EN\_1\_R} &
  \multicolumn{1}{c|}{H/4 * W/4} &
  \multicolumn{1}{c|}{8} &
  5 * 5 \\ \hline
\multicolumn{1}{c|}{2} &
  \multicolumn{1}{c|}{EN\_1 \& EN\_1\_R} &
  \multicolumn{1}{c|}{H/4 * W/4} &
  \multicolumn{1}{c|}{EN\_2} &
  \multicolumn{1}{c|}{H/8 * W/8} &
  \multicolumn{1}{c|}{32} &
  5 * 5 \\ \hline
\multicolumn{1}{c|}{2} &
  \multicolumn{1}{c|}{EN\_2} &
  \multicolumn{1}{c|}{H/8 * W/8} &
  \multicolumn{1}{c|}{EN\_2\_R} &
  \multicolumn{1}{c|}{H/8 * W/8} &
  \multicolumn{1}{c|}{16} &
  5 * 5 \\ \hline
\multicolumn{1}{c|}{2} &
  \multicolumn{1}{c|}{EN\_2 \& EN\_2\_R} &
  \multicolumn{1}{c|}{H/8 * W/8} &
  \multicolumn{1}{c|}{mu \& sigma} &
  \multicolumn{1}{c|}{H/8 * W/8} &
  \multicolumn{1}{c|}{2} &
  1 * 1 \\ \hline
\multicolumn{7}{c}{DECODER} \\ \hline
\multicolumn{1}{c|}{Layer} &
  \multicolumn{1}{c|}{Input} &
  \multicolumn{1}{c|}{Input Size} &
  \multicolumn{1}{c|}{Output} &
  \multicolumn{1}{c|}{Output size} &
  \multicolumn{1}{c|}{output kernel number} &
  Kernel Size \\ \hline
\multicolumn{1}{c|}{1} &
  \multicolumn{1}{c|}{sigma * gaussian + mu} &
  \multicolumn{1}{c|}{H/8 * W/8} &
  \multicolumn{1}{c|}{D\_1} &
  \multicolumn{1}{c|}{H/8 * W/8} &
  \multicolumn{1}{c|}{32} &
  1 * 1 \\ \hline
\multicolumn{1}{c|}{1} &
  \multicolumn{1}{c|}{D\_1 \& EN\_2 \& EN\_2\_R} &
  \multicolumn{1}{c|}{H/8 * W/8} &
  \multicolumn{1}{c|}{D\_1\_R} &
  \multicolumn{1}{c|}{H/8 * W/8} &
  \multicolumn{1}{c|}{16} &
  5 * 5 \\ \hline
\multicolumn{1}{c|}{2} &
  \multicolumn{1}{c|}{D\_1 \& EN\_2 \& EN\_2\_R \& D\_1\_R} &
  \multicolumn{1}{c|}{H/8 * W/8} &
  \multicolumn{1}{c|}{D\_2} &
  \multicolumn{1}{c|}{H/4 * W/4} &
  \multicolumn{1}{c|}{16} &
  5 * 5 \\ \hline
\multicolumn{1}{c|}{2} &
  \multicolumn{1}{c|}{D\_2 \& EN\_1 \& EN\_1\_R} &
  \multicolumn{1}{c|}{H/4 * W/4} &
  \multicolumn{1}{c|}{D\_2\_R} &
  \multicolumn{1}{c|}{H/4 * W/4} &
  \multicolumn{1}{c|}{8} &
  5 * 5 \\ \hline  
\multicolumn{1}{c|}{3} &
  \multicolumn{1}{c|}{D\_2 \& EN\_1 \& EN\_1\_R \& D\_2\_R} &
  \multicolumn{1}{c|}{H/4 * W/4} &
  \multicolumn{1}{c|}{D} &
  \multicolumn{1}{c|}{H/2 * W/2} &
  \multicolumn{1}{c|}{16} &
  5 * 5 \\ \hline
\multicolumn{7}{c}{UP-SAMPLING} \\ \hline
\multicolumn{1}{c|}{Layer} &
  \multicolumn{1}{c|}{Input} &
  \multicolumn{1}{c|}{Input Size} &
  \multicolumn{1}{c|}{Output} &
  \multicolumn{1}{c|}{Output size} &
  \multicolumn{1}{c|}{output kernel number} &
  Kernel Size \\ \hline
\multicolumn{1}{c|}{1} &
  \multicolumn{1}{c|}{D \& X\_0 \& FE\_1 \& FE\_2} &
  \multicolumn{1}{c|}{H/2 * W/2} &
  \multicolumn{1}{c|}{Y} &
  \multicolumn{1}{c|}{H * W} &
  \multicolumn{1}{c|}{1} &
  5 * 5 \\ \hline
\end{tabular}
\end{table*}
Figure~\ref{fig:structure} shows the structure of the proposed StereoVAE. It
consists of feature extraction, VAE, and upscaling modules; VAE can be
subdivided into encoder and decoder units.
Our StereoVAE receives two types of input images: one is the original left
image, and the other is the disparity map generated using traditional methods
in the previous step. Both are a quarter size of the original image.
The output is a high-resolution disparity map amplified by our network.
To improve the performance, we also employ skip connections and residual blocks
in the network.

Table~\ref{STRUCTURE OF STEREOVAE} shows the structure of our StereoVAE
network. $X\_0$ and $X\_1$ represent the low-resolution disparity map and the
grayscale left image as feature-extraction (FE) module inputs,
respectively. These two inputs are combined to extract the boundary
correspondence information between the disparity map and the left image because
disparity changes are usually drastic in these regions.
The {\em FE} module reduces the number of features in a
dataset by creating new features, which are represented by the existing ones, such as
boundary regions. These new features should then be able to summarize most of
the information contained in the original dataset. Since disparity maps usually
have fewer features than those general images have, our network simply employs two
convolutional layers each with 32 5$\times$5 convolutional kernels and a stride
of 1 to extract feature information from the low-resolution inputs.

To improve the learning and generalization ability of the network, we introduce
a VAE structure. The objective of the VAE module is to
reconstruct the features extracted by the {\em FE} module; the inputs {\em
  FE\_1} and {\em FE\_2} of the encoder are the outputs of the {\em FE}
module. The number of the outputs of the first layer is 16 with kernels
of 5$\times$5 and a stride of 2.
In the encoder module, the images are downsampled twice, along with two
residual blocks {\em R} to ensure the backward propagation of features. Each
{\em R} consists of a 5$\times$5 convolutional layer with a stride of 1.
The outputs of the encoder module second layer are downsampled to two
dimensions, {\em mu} and {\em sigma}, using a convolutional kernel of
1$\times$1. {\em mu} denotes the mean of a normal distribution, and {\em sigma}
denotes the variance logarithm of a normal distribution. The training
target of the VAE is to obtain a standard normal distribution whose mean is 0
and variance is 1. For both {\em mu} and {\em sigma}, their training objectives
are 0. However, the activation function used in this network is the {\em
  leakey\_relu}, which has a derivative of $0.1x$ for $x<0$ and $x$ for $x>0$,
and causes the network to learn at different rates for the positive and negative parts of
0. To eliminate the possible effects due to the difference in the learning speed, we
subtract 1 from {\em mu} and {\em sigma} to obtain the mean of the
normal distribution and the logarithm of the variance, as described in the following equation:
\begin{equation}
  \begin{aligned}
    gaussian = (mu - 1)  +  \varepsilon * exp^{(sigma-1)}.
  \end{aligned}
  \label{equation:Gaussian}
\end{equation}
Here, {\em mu} and {\em sigma} are obtained from the encoder module, $\varepsilon$
represents a standard normal distribution, and {\em Gaussian} is used as the
input of the decoder module.

The input of the decoder module is a normal distribution with the mean and variance
obtained from the encoder module, which is first increased to 32 dimensions using a
deconvolution layer with a 1$\times$1 matrix size and a stride of 1. Output
{\em D\_1} is used with {\em EN\_2} and {\em EN\_2\_R} to extract feature
information using a convolutional layer with a 5$\times$5 kernel. To avoid
overfitting, we employ skip connections to enhance the performance of
the proposed network. The number of feature maps in the first residual block is 16,
which is half of that in the first layer within the decoder module.
Similar to the encoder module, feature maps in the decoder module are
accompanied by two residual blocks and upsampled twice using two 5$\times$5 deconvolution
kernel and a stride of 2. After each upsampling, feature maps of
the same dimension within the encoder module are integrated and propagated
backward using the skip-connection mechanism. Finally, the final
high-resolution disparity map can be obtained using a separate upsampling
layer.

\subsubsection{Loss function}
The loss function of the StereoVAE consists of the following two parts: 1) the difference
between the high-resolution disparity map and the ground truth; 2) the
difference between the output of the encoder module and its standard normal
distribution. Here,
since we normalize the image data in the first step, the value of the two loss
functions must be expanded by a factor of 256 to ensure that the results are the same
as those calculated in an 8-bit format. Then, the final loss function can be
defined as follows:
\begin{equation}
  Loss_{total} = 256* (Loss_{1} + Loss_{2}).
  \label{equ:Loss func}
\end{equation}
For $Loss_1$, we use the absolute error between the output and the ground truth
and set the weight of the non-occluded points to twice that of the occluded
points. Then, the specific loss $Loss_{1}$ can be defined as follows:
\begin{equation}
  \begin{aligned}
    &Loss_{noc} = |gt_{noc} - Y|,\\
    &Loss_{occ} = |gt_{occ} - Y|,\\
    &Loss_1 = Loss_{noc} + Loss_{occ}.
  \end{aligned}
  \label{equation:LOSS1}
\end{equation} 
Here, $Y$ is the output of the StereoVAE network, $gt_{noc}$ is the ground truth without considering occlusion points, and $gt_{occ}$
represents is the ground truth when considering occlusion points. Furthermore, $Loss_2$ can be expressed as
follows:
\begin{equation}
  \begin{aligned}
    Loss_{2} = 0.5 &* \{(mu - 1)^2 - 1 
    \\+ &exp^{(sigma-1)^2} 
    \\- &log(exp^{(sigma-1)^2})\}
  \end{aligned}
  \label{equation:LOSS2}
\end{equation}
where {\em mu} and {\em sigma} represent the outputs of the encoder module. As
the network iterates, both {\em mu} and {\em sigma} are expected to
converge to 1.
\begin{figure*}
\centering \includegraphics[width=6.5in]{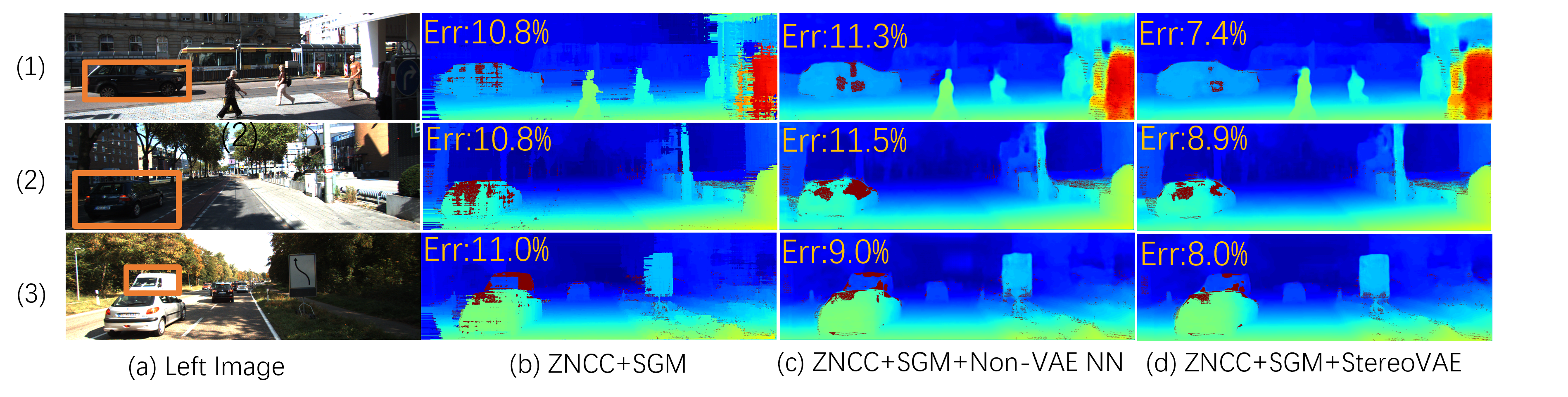}
\caption{Accuracy Comparison of StereoVAE using Different Methods. (1) Textureless region. (2) Specular region. (3) Brightness region.}
\label{fig:comparsion1}
\end{figure*}

\section{Experiments}
\label{sec:4}
In this section, we initially introduce the dataset and training conditions employed
in our experiment, and then we present the experimental results to demonstrate the efficiency of the proposed system regarding the following aspects:
\begin{itemize}
\item comparison with traditional and non-VAE learning-based methods in terms of
  accuracy and processing speed;
\item robustness testing of our StereoVAE on both GPUs (Jetson Tx2 and Jetson
  AGX Xavier) by refining the disparity maps generated using different traditional
  methods;
\item performance comparison with existing embedded systems.
\end{itemize}

\subsection{Preparation}
The initial quarter size of disparity images was obtained by combining ZNCC and SGM with a half disparity range of 64. These two
traditional methods were fully optimized using the CUDA platform to generate a coarse
disparity map. The StereoVAE network was trained using the KITTI 2015
dataset, which is commonly used in autonomous driving research. The KITTI 2015
dataset contains 200 training image pairs and 200 testing image pairs. 160 of them were used for training and 40 for validation, respectively. The Adam optimizer was selected as the network optimizer, where $\beta_1$ was set to 0.99 and $\beta_2$ was set to 0.999. The
initial learning rate was 0.0005. 1000 epochs were trained, and the
learning rate was reduced by a factor of 3\% every 10 epochs. The batch size was set to 1. Our
StereoVAE has no constraints other than the loss function and an image cropping
step.

\subsection{Performance Evaluation}
Traditional and non-VAE learning-based methods were compared with
our method in terms of processing speed and accuracy. Table~\ref{table2}
shows the running speed of these three methods on both GPUs (Jetson TX2 and
Jetson AGX Xavier). It is observed that the traditional method (ZNCC+SGM)
maintains its advantage in terms of processing speed (28 fps and 79 fps), which
is roughly twice faster than that of our StereoVAE (12 fps and 30 fps). However,
our system can achieve real-time processing on the Jetson AGX Xavier
GPU. The non-VAE learning-based method is slightly faster (14fps and 36 fps)
than our StereoVAE, which means that the VAE structure does
not spend too many computational resources.
\begin{table}[h]
\centering
\caption{StereoVAE EVALUATION}
\label{table2}
\begin{tabular}{l|l|l}
 \hline
 Method & Running time (Tx2)& Running time (AGX)  \\ 
  \hline
ZNCC+SGM & 35.71 ms & 12.64 ms    \\ 
  \hline
Non-VAE NN & 72.1 ms & 27.9 ms  \\ 
  \hline
StereoVAE &  84.18 ms & 29.88 ms  \\ 
  \hline
\end{tabular}
\end{table}
Figure~\ref{fig:comparsion1} shows the accuracy performance of these three
methods under different image pairs. Regarding the error rate, ZNCC + SGM +
StereoVAE $>$ ZNCC + SGM + non-VAE NN $>$ ZNCC + SGM fully demonstrates that our
method is influential in the optimization. Since the traditional
methods selected are based on pixel matching, the disparity maps they generated
lack detailed optimization and contain much noise. In this aspect, learning-based
methods are superior. Especially in the textureless,
specular, and brightness regions (shown in orange boxes), traditional methods cannot produce satisfactory results due to the lack of matching
features. In contrast, our VAE method exhibits a broader latent space expression
and prediction ability.
\begin{table*}
  \caption{COMPARISON AMONG VARIOUS EMBEDDED STEREO-MATCHING SYSTEMS THAT EMPLOY THE KITTI 2015 DATASET}
\begin{center}
\begin{tabular}{c|cc|cc|cc|c|c}
  \hline
  \cline{2-9}
  \textbf{Error rate (\%)} &\multicolumn{2}{c|}{\textbf{D1-bg}}&\multicolumn{2}{c|}{\textbf{D1-fg}}&\multicolumn{2}{c|}{\textbf{D1-all}}&\multicolumn{2}{c}{\textbf{Speed}}\\
  \cline{2-9}
  &\textbf{All/All}&\textbf{Noc/All}&\textbf{All/All}&\textbf{Noc/All}&\textbf{All/All}&\textbf{Noc/All}&\textbf{GPU}&\textbf{fps}\\
  \hline
  StereoNet\cite{c21} & 4.3 & ---  & 7.45  & --- & 4.83 & ---& Tx2 & 1   \\
  MADNet\cite{c22} & 3.75 & 3.45  & 9.20  & 8.41 & 4.66 & 4.27& Tx2 & 4  \\
  RTS2Net\cite{c7} & 3.09 & ---  & 5.91  & --- & 3.56 & ---& Tx2 & 6  \\
  RTSMNet\cite{c8} & 3.44 & 3.21  & 6.08  & 5.39 & 3.88 & 3.57& AGX & 11 \\
  GPUBNN\cite{c9}  & --- & 3.5  & ---  & --- & --- & 4.57& AGX & 10 \\
  Res2tAC\cite{c3}  & 6.27 & 5.14  & 16.07  & 14.29 & 7.9 & 6.65& AGX & 15  \\
  DWARF\cite{c23}  & 3.2 & 2.95 & \textbf{3.94}  & \textbf{3.66} & 3.33 & 3.07& Tx2 & 1  \\
  AnyNet\cite{c4} & 6.32 & 6.01 & 13.93 & 13.11 & 7.59 & 7.18& AGX & 26\\
  StereoDNN\cite{c24} & \textbf{2.7} & \textbf{2.1} & 6.0 & 4.5 & \textbf{3.2} & \textbf{2.5}& Tx2 & 1\\
  \hline
  \textbf{Our StereoVAE} & 4.71 & 4.38 & 7.88 & 6.48 & 5.24 & 4.73& AGX & \textbf{30} \\
  \hline
\end{tabular}
\label{table3}
\end{center}
\end{table*}
\subsection{Robustness}
To evaluate the robustness of our StereoVAE, we compared its optimization
performance using four combinations of traditional methods: S1(Census+DT),
S2(Census+SGM), S3(ZNCC+DT), and S4(ZNCC+SGM). Figures~\ref{fig:e4} and
\ref{fig:e5} compare the accuracy of different methods that employ the KITTI 2015
validation and testing datasets, respectively. $Mini+Linear$ refers to directly using
traditional methods to perform the stereo matching on small image pairs with a
linear scaling. $Original$ represents the results obtained by applying direct matching on large
image pairs. $Mini+StereoVAE$ represents the results obtained by applying the proposed
method. Our StereoVAE improves the matching accuracy in all cases. The largest error rate reduction achieved by StereoVAE was 5.82\%, which clearly demonstrates the high robustness level of the proposed method. 

\begin{figure}
\begin{tikzpicture}[scale=0.8]
\begin{axis}[
    ybar,
    bar width=11pt,
    width=4in,
    height=2.5in,
    enlargelimits=0.15,
    legend style={at={(0.5,-0.15)},
      anchor=north,legend columns=-1},
    ylabel={Error Rate (\%)},
    symbolic x coords={S1-based,S2-based,S3-based,S4-based},
    xtick=data,
    nodes near coords,
    every node near coord/.append style={font=\scriptsize},
    nodes near coords align={vertical},
    ]
\addplot  [cyan!40!black,fill=cyan!40!white] coordinates {(S1-based,17.6) (S2-based,9.73) (S3-based,10.95) (S4-based,9.56)}; 
\addplot [yellow!40!black,fill=yellow!40!white]  coordinates {(S1-based,13.2) (S2-based,7.09) (S3-based,8.8) (S4-based,6.4)};
\addplot [red!40!black,fill=red!40!white] coordinates {(S1-based,11.78) (S2-based,4.57) (S3-based,7.56)(S4-based,4.64)};
\legend{Mini+Linear, Original, Mini+StereoVAE}
\end{axis}
\end{tikzpicture}
\caption{Accuracy comparison among various systems that employ the KITTI 2015 validation dataset. Lower values mean better results.}
\label{fig:e4}
\end{figure}
\begin{figure}
\begin{tikzpicture}[scale=0.8]
\begin{axis}[
    ybar,
    bar width=11pt,
    width=4in,
    height=2.5in,
    enlargelimits=0.15,
    legend style={at={(0.5,-0.15)},
      anchor=north,legend columns=-1},
    ylabel={Error Rate (\%)},
    symbolic x coords={S1-based,S2-based,S3-based,S4-based},
    xtick=data,
    nodes near coords,
    every node near coord/.append style={font=\scriptsize},
    nodes near coords align={vertical},
    ]
\addplot [cyan!40!black,fill=cyan!40!white] coordinates {(S1-based,23.8) (S2-based,8.52) (S3-based,10.35) (S4-based,7.47)}; 
\addplot [yellow!40!black,fill=yellow!40!white] coordinates {(S1-based,14.5) (S2-based,7.03) (S3-based,8.26) (S4-based,6.24)};
\addplot [red!40!black,fill=red!40!white] coordinates {(S1-based,8.68) (S2-based,5.41) (S3-based,6.43) (S4-based,4.73)};
\legend{Mini+Linear, Original, Mini+StereoVAE}
\end{axis}
\end{tikzpicture}
\caption{Accuracy comparison among various systems that employ the KITTI 2015 testing dataset. Lower values mean better results.}
\label{fig:e5}
\end{figure}

\subsection{Comparison With Other Systems}
Table~\ref{table3} shows the comparison results of our system with other existing
embedded stereo-vision systems. All systems were implemented on embedded GPUs,
and achieved good performance. StereoDNN~\cite{c24} exhibits the highest matching
accuracy with only a 2.5\% error rate. However, its processing speed is lower than 1 fps and thus it cannot meet the real-time requirements of embedded
applications. On the other hand, our system exhibits the best processing speed
in the embedded stereo-vision system list, while the accuracy is higher
than that of the fast systems AnyNet~\cite{c4} and Res2tAC~\cite{c3}, demonstrating that our system achieves a good balance regarding the embedded stereo-vision
system performance.

\section{CONCLUSIONS}
\label{sec:5}
In this study, we proposed a high-performance stereo-matching system implemented on a Jetson AGX Xavier GPU. 
The proposed hybrid structure includes of 1) the generation of a coarse disparity
map using traditional methods and 2) a VAE-based neural network to upscale
and refine the disparity map. Extensive experiments on the KITTI
2015 dataset, our system exhibits high processing speed and accuracy
performance, indicating that our system combines the advantages of high processing speed of traditional methods and high accuracy of
learning-based methods. However, there is still a gap in terms of accuracy compared with
the current state-of-the-art methods, mainly because the proposed lightweight network lacks
the required convolution kernels.

We plan to further compress our StereoVAE network through quantization to
achieve increased processing speed and low memory usage, which can
facilitate its application to other hardware platforms.

\addtolength{\textheight}{-12cm}   



\section*{ACKNOWLEDGMENT}
This work was partly supported by JSPS KAKENHI, Grant No. 21K17868, and JST CREST, Grant No. JPMJCR22M2.


\end{document}